\documentclass[
]{ceurart}

\usepackage{listings}
\usepackage[utf8]{inputenc}
\usepackage{alltt}
\usepackage{todonotes}
\usepackage{caption}
\usepackage{subcaption}
\usepackage{float}
\usepackage{fancyvrb}

\newfloat{lstfloat}{htbp}{lop}
\floatname{lstfloat}{Listing}

\begin{document}
\copyrightyear{2022}
\copyrightclause{Copyright for this paper by its authors.
  Use permitted under Creative Commons License Attribution 4.0
  International (CC BY 4.0).}

\conference{NeSy 22: 16th International Workshop on
Neural-Symbolic Learning and Reasoning, 28-30 September, London, UK}

\title{ESC-Rules: Explainable, Semantically Constrained Rule Sets}


\author[1]{Martin Glauer}[%
orcid=0000-0001-6772-1943,
email=martin.glauer@ovgu.de,
]
\cormark[1]
\address[1]{Institute for Intelligent Interacting Systems, Otto-von-Guericke University Magdeburg, Germany}

\author[2]{Robert West}[%
orcid=0000-0001-6398-0921,
]

\author[3]{Susan Michie}[%
orcid=0000-0003-0063-6378,
]

\author[1,3]{Janna Hastings}[%
orcid=0000-0002-3469-4923,
email=j.hastings@ucl.ac.uk,
url=https://jannahastings.github.io/,
]
\cormark[1]
\address[2]{Department of Behavioural Science and Health, University College London, UK}
\address[3]{Department of Clinical, Educational and Health Psychology, University College London, UK}

\cortext[1]{Corresponding author.}

\begin{abstract}
    We describe a novel approach to explainable prediction of a continuous variable based on learning fuzzy weighted rules. Our model trains a set of weighted rules to maximise prediction accuracy and minimise an ontology-based 'semantic loss' function including user-specified constraints on the rules that should be learned in order to maximise the explainability of the resulting rule set from a user perspective. This system fuses quantitative sub-symbolic learning with symbolic learning and constraints based on domain knowledge. We illustrate our system on a case study in predicting the outcomes of behavioural interventions for smoking cessation, and show that it outperforms other interpretable approaches, achieving performance close to that of a deep learning model, while offering transparent explainability that is an essential requirement for decision-makers in the health domain. 
\end{abstract}

\begin{keywords}
  explainable machine learning \sep
  rules learning \sep
  semantic loss \sep
  evidence synthesis \sep
  behaviour change
\end{keywords}

\maketitle

\section{Introduction}
The rate evidence is generated outstrips the rate at which it can be synthesised, necessitating automated approaches \cite{elliott_decision_2021}.
The Human Behaviour-Change Project (HBCP) is an interdisciplinary collaboration between behavioural scientists, computer scientists and information systems architects that aims to build an end-to-end automated system for evidence synthesis in behavioural science \cite{michie_human_2020}. For this objective, explanations of the predictions are as important as the accuracy of the system, since the intended users are practitioners and policy-makers who will use the insights gained from the evidence in order to make recommendations thus require suitable transparency and accountability \cite{kelly_key_2019}. 

Deep neural networks typically operate as black boxes without giving intrinsic insights into why specific predictions have been made.\footnote{Although, methods to give explanations for such networks are advancing \cite{moradi_post-hoc_2021,madsen_post-hoc_2021}.} Thus, there is a need for ``glass-box'' explainable machine learning frameworks for making predictions and recommendations that can transparently provide complete explanations in a form that matches the semantic expectations of the users \cite{holzinger_information_2022,holzinger_xxai_2022}. 


We aimed to develop an explainable system for the prediction of behaviour change intervention outcomes, based on a corpus of annotated literature together with features from an ontology - the Behaviour Change Intervention ontology \cite{michie_representation_2020} - and their logical relationships. Straightforward application of semantic approaches is not well suited for the quantitative task of predicting intervention outcomes, and moreover traditional symbolic learning approaches such as rules or decision trees lead to explanations that are overly complex and not ranked by their quantitative impact on the outcome variable. A deep neural network approach had better quantitative performance, but was not acceptable for our users due to the lack of transparency and explainability. Thus, we aimed to develop a `best of both worlds' hybrid predictive approach that combined aspects of the symbolic and neural approaches. 

Rules-based systems are inherently explainable because the features appear transparently in the rules. Our approach builds on systems that are able to learn rules from data. One of the earliest learning rule neural network systems was the Knowledge Based Artificial Neural Network - KBANN \cite{towell_knowledge-based_1994}. This approach translates domain knowledge into rules which are encoded into the structure of a neural network, for which weights are then learned. 
The approach that we developed has furthermore been inspired by traditional `neuro-fuzzy' systems. It is, in particular, based on the principles of Tagaki-Sugeno controllers \cite{takagi1985fuzzy}. These controllers have been developed to account for the fact that expert knowledge, albeit valuable, is often too vague to be turned into rigid logical rules, thus necessitating fuzzy and weighted approaches to rules learning.

\section{Explainable Semantically Constrained Rule Sets}

We describe our approach in terms of feature preparation, architecture and optimisation, and semantic penalties. The system is implemented in Python using PyTorch. Source code is available at \texttt{https://github.com/HumanBehaviourChangeProject/semantic-prediction}.

\subsection{Feature preparation}
A precondition for our rules-based approach is that all input data features are binarized. Thus, categorical variables in the dataset are exploded into separate columns per value, and continuous (quantitative) variables are binarized by selecting ranges using one of several different approaches depending on the meanings of the values: 
\begin{itemize}
    \item Separation into meaningful semantic categories, e.g. for our case study we transform mean age values into child, young adult, older adult, and elderly, delineated with a fuzzy membership operator since the boundaries between categories are not rigid. 
    \item Fixed-width categories, delineated with a fuzzy membership operator. For example, we divide number of times tobacco smoked into groups of width 5 corresponding to <5, <10, <15, ... <50. 
    Note that this formulation creates an ordering because if the value is e.g. 6, then all of <10, <15, ..., <50 will be set on, and if the value is 46, only <50 will be set on. 
    \item Categories selected based on quantiles in the dataset (i.e. a fixed proportion of the available data in each grouping, rather than fixed range of values), again using the <x formulation to maintain ordering. 
    \item Fixed numeric values, for flagging exact values with a specific meaning, for example 100\% female within the population percentage female continuous variable. 
    \item Data-driven clustering can be used to determine clusters of data associated with specific value ranges. 
\end{itemize}



\subsection{Architecture and Optimisation}

The general design of the architecture was inspired by the construction of Takagi-Sugeno controllers. But, while that approach assumes that a set of rules is given as a prior, the goal of our system is to automatically derive the rules from the dataset. We assume that rules are represented as conjunctive formulae of all features in the dataset and their negations. A weight is then attached to each rule and each feature in the rule as well as their negated counterparts:
\begin{equation}
(w_{i,1}, A_{1}) \wedge \cdots \wedge (w_{i,n},A_{n}) \wedge (w_{i,n+1}, \neg A_{1}) \wedge \cdots \wedge (w_{i,2n},\neg A_{n}) \Rightarrow r_i
\end{equation}

A weight $w_{i,j}\in \mathbb{R}$ is attached each literal $A_j$. The sigmoid $\sigma(w_{i,j})$ of these weights denotes the degree to which literal $A_j$ impacts the $i$-th rule. Each rule is also attached with a rule weight $r_i \in \mathbb{R}$ that denotes the impact of this rule on the prediction. For the sake of readability, we will further refer to the literals $A_1,...,A_n,\neg A_1,...,\neg A_n$ (i.e. features and negated features) as just $B_1,...,B_{2n}$. The basic idea of our approach is to hide the parts of the conjunction that are not relevant to a particular rule. That is, variables with a small weight should not influence the rule as much as ones with larger weights. Given a fuzzy evaluation function that assigns a fuzzy membership value to each feature $\mu: \{B_1,...,B_{2n}\} \rightarrow [0,1]$, we therefore calculate the weighted conjunctive contribution of a literal $B_j$ in rule $i$ as a linear combination between $\mu(B_{j})$ and 1. This allows us to finally define the fit of a rule as $\text{fit}_i = \max_{j=1}^{2n}(\sigma(w_{i,j})\mu(B_{j}) + (1- \sigma(w_{i,j})))$. Similar to the Takagi-Sugeno controller approach, this fit is an indicator of how much the right-hand side of the rule should impact the final prediction of the system. Our system uses this fit as a simple scaling factor $y_\text{pred} = \sum^m_{i=1} \text{fit}_i \cdot r_i$ to aggregate predictions from multiple rules.

\subsection{Semantic penalties}
While the system described so far does learn rules, those rules are not meaningfully explainable. This is mainly due to two effects: A) The sigmoids of the weights often take arbitrary values from [0,1]. B) The left-hand sides of the rules are often very long, which makes it difficult to analyse the system easily. To address these points, we introduce additional regularisation terms that penalise such behaviour. The first one $\lambda_\text{long} = \sum^m_{i=1} \max \left( \sum_{j=1}^n\sigma(w_{i,j}), \theta \right)$ penalises rules with length exceeding the threshold $\theta$. The second term $\lambda_\text{fuzzy} = \sum^m_{i=1}  \sum_{j=1}^n (1-\sigma(w_{i,j}))\sigma(w_{i,j})$ uses a quadratic equation for non-crisp literals. This penalty is only non-zero for those feature weights that are non-crisp. 

These penalties improve the learned rules. However, the rules learned are still somewhat arbitrary and lack semantic coherence. Thus, as a final measure we add semantic ontology-based penalties that aim  to increase the coherence and sense of the rules that are learned using the background domain knowledge as embodied in an ontology. For our use case we derive the rules from the Behaviour Change Intervention Ontology \cite{michie_representation_2020}, using several semantic features. 

\begin{description}

\item[Implied features] The ontology contains a class hierarchy, which is used in two different ways. First, to pre-complete the data table: if a lower level feature is on, we ensure that the higher level feature that subsumes it is also set on in the input dataset. Hierarchical implications and other implications deriving from other types of relationships between entities in the ontology are also used to reduce rule length and enhance rule interpretability. Intuitively, a rule that has two features, such as '$\text{Somatic}(X) \wedge \text{PharmacologicalSupport}(X) \Rightarrow r$', where one feature implies the other, can be reduced to a rule with one feature, e.g. '$\text{PharmacologicalSupport}(X) \Rightarrow r$', because of the implication given by the ontological relationship between the entities, as every intervention with pharmacological support is also one with a somatic delivery mode. This results in shorter, more readable rules. Therefore, we introduce an additional penalty $\lambda_{\text{implied}} = \sum^m_{i=1}  \sum_{(j,j') \in \mathcal{I}} \min(\sigma(w_{i,j}),\sigma(w_{i,j'}))$ for the co-occurence of implied features in each rule.

\item[Mutually exclusive features]

The ontology also contains axioms regarding negative dependencies between features. For example, an intervention that has the feature `buproprion’ (a pharmaceutical that is administered in form of a pill) cannot have the feature `not pill’. These dependencies can, for instance, be expressed as disjointness axioms. Rules that include features that contradict each other semantically are penalised using the same mechanism that has been employed for the implications. This penalty $\lambda_{\text{exclusive}}$ is also applied to logically contradictory features within rules, e.g. $A \wedge \neg A \rightarrow r$.

\end{description}

%
%

These penalties ensure the logical soundness and usability of the final rules. Of course, it would be possible to use purely symbolic techniques to achieve similar behaviour. Using penalties does however have the added benefit that the neural network can learn which parts of a conflicting rule can be dropped with minimal impact on predictive performance, and thereby find a better fit for the data.

Additional penalties increase the complexity of the training task considerably, as a multitude of different goals have to be optimised. We therefore use an additional scaling factor $\alpha$ that fades these penalties in as the training progresses to later epochs. The final loss function is therefore calculated using the binary cross entropy (bce):
\[
\text{loss}(y_\text{pred}, y_\text{target}) = \text{bce}(y_\text{pred}, y_\text{target}) + \alpha \left(\lambda_\text{long}  + \lambda_\text{fuzzy} + \lambda_\text{impl} + \lambda_{\text{exclusive}}\right)
\]
The resulting system learns human-readable rules from data. These rules are constrained to be simple and semantically meaningful (and explainable) through the semantic regularisation terms. 

\section{Evaluation}
\subsection{Dataset}

The dataset consists of features extracted from randomised controlled trials of smoking cessation interventions by manual annotation as described in \cite{bonin_hbcp_2020}. Features - annotated using ontology classes - cover several different aspects of intervention trial reports, including population attributes (e.g. mean age), delivery attributes such as who delivered it (e.g. nurse) and how it was delivered (e.g. face to face), and types of intervention (e.g. pharmacological support or goal setting). In the corpus, these features are associated with the text specifying the presence of the feature and any associated value, as well as the wider surrounding contextual text, and the whole composite is encoded in JSON. From the JSON, additional processing is used to extract tabular data with a column for each feature and a row for each intervention arm, and some additional data cleaning ensures that numeric values are appropriately parsed and that units are regularised. 
The resulting table has 77 feature columns and 1198 rows corresponding to intervention arms. 


\subsection{Model and Results}
We evaluated our system against four other approaches. Firstly, we compare to the model that was previously developed within the HBCP \cite{ganguly2021outcome}, which we call `NLPG' (short for NLP+Graph, as it learns from a combination of the textual feature contexts and the annotations encoded as a graph). This model was the previous best approach for this problem and has been applied as-is as it works directly from the data encoded in the JSON corpus. The remainder of the approaches work from the tabular feature dataset described above, which includes additional data cleaning steps such as ensuring consistent units for numeric features. Thus, for additional points of comparison, we also applied three different models from different families of machine learning approach. 

\begin{description}
    \item[decision trees] A random forest of 100 trees with a maximal depth of 4 layers and 7 leaf nodes.
    \item[deep] A feed-forward neural network with 77 input neurons and 3 hidden layers with 154, 77 and 38 neurons,  trained for 100 epochs using the Adam optimizer. 
    \item[mixed-linear] As baseline we use a mixed-effect linear regression model with a random effect for study and fixed effects for all other variables. 
\end{description}

\begin{figure}
\centering
    \begin{subfigure}{0.32\textwidth}
        \centering
        \includegraphics[width=\textwidth]{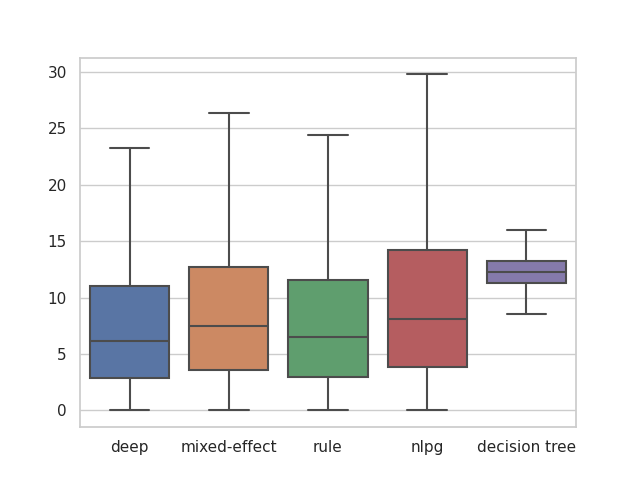}
    \end{subfigure}
    \begin{subfigure}{0.32\textwidth}
        \centering
        \includegraphics[width=\textwidth]{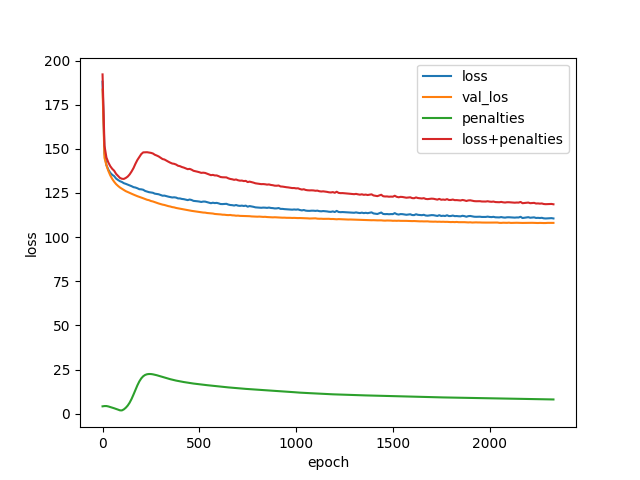}
    \end{subfigure}
    \begin{subfigure}{0.32\textwidth}
        \centering
        \includegraphics[width=\textwidth]{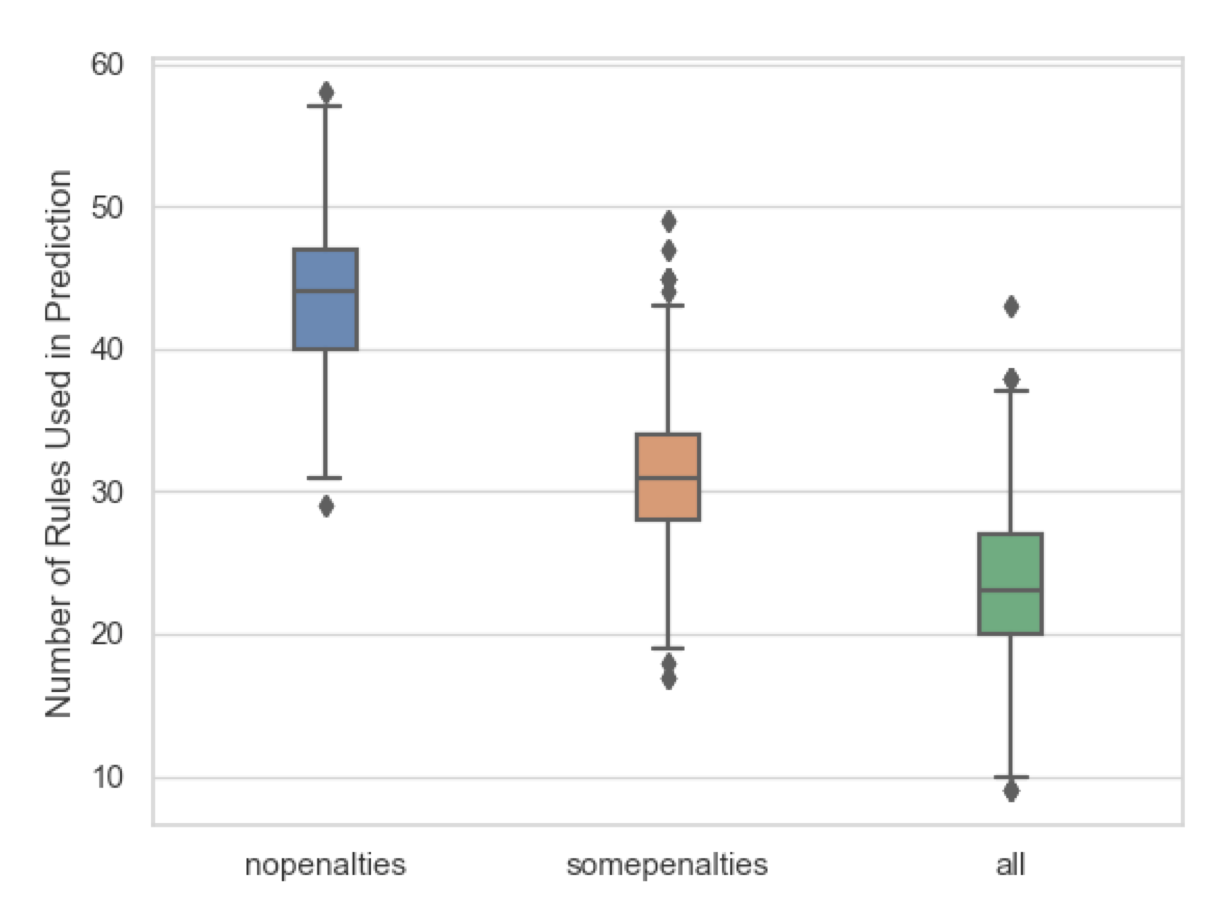}
    \end{subfigure}
    \caption{Comparisons of the results of different models. Left: Box plot showing the distribution and mean of the absolute errors in each model. Centre: Training/validation loss and penalties. Penalties are faded it at the beginning of the training. Right: Numbers of rules used in predictions with no, some or all penalties.}
    \label{fig:err_plots}
\end{figure}

Our model was initialised with a set of 100 rules. Singleton rules for each (non-negated) feature were initialised based on the weights from a simple linear regression, and the remaining 23 were initialised randomly using two uniform distributions: $\mathcal{U}(-1,1)$ for feature weights and $\mathcal{U}(-10,10)$ for rule weights. The resulting model was trained for at least 300 epochs, after which the training was stopped as soon as the loss (including penalties) did not further improve on the validation set. The final evaluation was then conducted on a test set that had not been seen before. 

Figure \ref{fig:err_plots} shows the performances of each model in terms of the absolute errors of the predictions on the unseen test set data. Our model outperforms the original approach for prediction, as well as decision trees and the mixed linear model. It is still slightly outperformed by the pure deep neural network, however, that approach is not explainable. Additional evaluations have shown that training on a smaller subset of the dataset leads to less robust results.

The final state of our system can be used to extract human-readable rules that apply for each prediction. Some examples of rules that are used in predictions are illustrated in Listing \ref{lst:rules}. 
\begin{lstfloat}
  \centering

\begin{BVerbatim}
"Pharmaceutical company competing interest"[1.0] 
  => raise predicted outcome by 0.2623 (fit: 0.9999)

"1.2 Problem solving"[1.0]
  => raise predicted outcome by 3.2550 (fit: 0.9999)

"Mean age (adult)"[1.0] & "doctor"[0.46218] & "Pill"[0.3893] 
  => raise predicted outcome by 4.6709 (fit: 0.5382)

"doctor"[0.4442] & "Biochemical verification"[1.0]
  => lower predicted outcome by 2.7093 (fit: 0.5563)

"aggregate patient role"[0.6734] & "Abstinence: Continuous"[1.0] 
  & not "1.4 Action planning"[1.0] & not "Somatic"[0.7504] 
  & not "Proportion identifying as female gender" (<= 35)[0.7507]
  => lower predicted outcome by 2.9024 (fit: 0.3272)
\end{BVerbatim}
\caption{Excerpt of a learned rule base that has been applied to a set of input values. Feature names are in quotations, followed by their attached weight. The denoted increases and decreases are already scaled according to the fit of the corresponding rule.}
\label{lst:rules}
\end{lstfloat}

As can be seen, the rules that are learned vary in length from simple weights for individual features, to complex interactions between features, and vary in their impact on the outcome value, which may be positive or negative. 

The rules and their fits are completely transparent,  allowing experts to inspect the functioning of the system and add additional semantic constraints where needed to improve the overall performance of the system, leading to an iterative process which may also have the benefit of improving the associated ontology from which the semantic constraints are drawn.


\section{Related Work}

Our work is related to several ongoing research areas within neuro-symbolic computing, which we discuss in turn. 

\subsection{Learning fuzzy logic operations in deep neural networks}

Our approach has aspects of a fuzzy inference system, which are less general than neural networks but more explainable. Most fuzzy neural systems are restricted to just a few logical operations, usually \texttt{and} and \texttt{or}. They present an activation function that can learn additional logical operations and allow learning of complex logical expressions with accuracy comparable to a standard deep neural network with \texttt{tanh} activation functions. A deep learning architecture has been proposed for learning fuzzy logic expressions in \cite{godfrey_parameterized_2017}. More generally, Logic Tensor Networks \cite{serafini2016logic} are able to represent arbitrary first-order formulae and learn the semantic interpretation of constants, predicate and function symbols during training by minimising a specific semantic loss function. The semantics for logical connectives are based on existing fuzzy semantics - usually based on {\L}ukasiewicz logic. Another notable approach to rule learning are differentiable logic machines\cite{zimmer_dlm_2021}. These networks allow the expression and learning of logical formulae and use a hiding mechanism similar to the one presented in this work. They are however more suitable for purely symbolic applications and less for the regression task that was our use-case.

\subsection{Logically constraining deep neural networks}

There is a significant body of work on constraining neural networks with decision rules (reviewed in \cite{okajima_deep_2019}) or with logical constraints (reviewed in \cite{giunchiglia_deep_2022}). A recent addition to this family of approaches is Deep Neural Networks with Controllable Rule Representations (DeepCTRL) \cite{seo_controlling_2021}, an approach that incorporates a rule encoder into a deep neural network model together with a rule-based objective. It allows specification of rules for inputs and outputs, with rule `strength' adjustable at inference time via a parameter (not requiring retraining). 
The rules constrain the model search space to reduce under-specification and improve generalisability, which is similar to the effect of the semantic ontology-based constraints in our approach. However, their rules are not based on an ontology thus must be manually specified.  


\subsection{Learning weighted rule sets}

The RuleFit algorithm \cite{friedman_predictive_2008} learns sparse linear models that include automatically detected interaction effects in the form of rules. 
Similarly to our approach, in this approach base variables take values in (0,1) with categorical variables one-hot encoded and numerical variables discretised into ranges. However, this approach uses pre-specified rule lengths rather than optimising the rule length together with the rule weights as we do. In RuleFit, the sizes of the rules are governed by a random variable selected from an exponential distribution so that smaller rules are favoured. Thus, most of the rules will be simple capturing main effects while some will be larger capturing interaction effects. Moreover, it does not incorporate semantic constraints on the rules as our system does. 

A two-layer neural network for learning rules for binary classification rather than regression, Decision Rules-net (DR-net) \cite{qiao_learning_2021} includes a first layer consisting of a set of neurons that each map to decision rules, and a second layer that performs a disjunctive combination of the rules from the first layer. Input features are binarized in a similar fashion to our system, and then the first layer operates as a conjunction over features in the input using a binary step activation function. This operation is not differentiable, thus it is approximated with the straight-through estimator using gradient clipping. Learnable weights are associated with each decision rule in the first layer of the network, with positive weights corresponding to a positive association of the input feature, negative weights with a negative association, and a zero weight corresponding to exclusion of the input feature. Stochastic gradient descent is used for training. Sparsity-based regularisation implements a tradeoff between accuracy of predictions and simplicity of explanations in terms of the length of the rules. However, they include no additional semantic penalties. 




\section{Conclusions}

We presented a system for rule learning on sparse data for a regression problem, and evaluated the performance on a use-case from the domain of behaviour change interventions. Our system out-performed other explainable methods, and was not far from the predictive power of a black-box deep neural network. Our system is able to generate human-readable rules that can be used to easily explain predictions. Moreover, it is possible to enhance the system through semantic constraints that can either be defined by experts or extracted from an ontology. This ensures that the system is able to generate rules that not only fit the dataset but also reflect domain knowledge, are semantically correct, explainable, and do not contain unnecessarily complex (and likely overfitted) conditions.

In the future, we aim to conduct a more in-depth evaluation study with users and domain experts, in particular with respect to the explainability of the rules. One particular aspect which we would like to explore more deeply with the domain experts is the explainability associated with the negated features, i.e. the explanatory value of \textit{absences} of features, where these appear in rules. We will also conduct a more in-depth comparison to newer neuro-symboli approaches. Another aspect we plan to explore is whether there is a distinction or preference among domain experts for rules that contain interaction terms from different semantic sub-domains (e.g. intervention, population, setting) or those that show interactions within a semantic domain (e.g. different types of intervention in combination). 

We also plan to apply the system on different datasets drawn from a wider range of behavioural domains, including physical activity. 

\begin{acknowledgments}
  Thanks to all members of the Human Behaviour-Change Project team and in particular to Alison Wright and James Thomas. The HBCP is funded by a Wellcome Trust collaborative award led by Susan Michie.
\end{acknowledgments}

\bibliography{bibliography} 

\end{document}